\documentclass[letterpaper, 10 pt, conference]{ieeeconf}  

\IEEEoverridecommandlockouts                              

\usepackage{amsmath}
\usepackage{amssymb}
\usepackage{graphicx}
\usepackage{booktabs}
\usepackage{tabularx}
\usepackage[table]{xcolor}
\usepackage{capt-of}
\usepackage{placeins}
\usepackage[nocompress]{cite}
\newsavebox{\rafalignmenttable}
\newlength{\rafablationwidth}

\title{\LARGE \bf
RAF-VLA: Representation Alignment with the Future \\ for End-to-End Autonomous Driving
}
\usepackage{pifont}

\author{Dogun Kim$^{1}$, Yongjae Lee$^{1}$, Joonhee Lim$^{2}$, Yeina Lee$^{2}$, Junhyeok Park$^{1}$, Moogeun Park$^{2}$, Dongsuk Kum\textsuperscript{1,\ding{41}}
\thanks{$^{1}$Graduate School of Mobility, Korea Advanced Institute of Science \& Technology (KAIST), Daejeon, Korea {\ttfamily (dogun.kim@kaist.ac.kr)}}%
\thanks{$^{2}$Robotics Program, Korea Advanced Institute of Science \& Technology (KAIST), Daejeon, Korea \ding{41}: Corresponding author.}%
}

\IEEEaftertitletext{%
  \vspace{-0.6em}%
  \noindent\begin{minipage}{\textwidth}%
    \centering
    \includegraphics[width=\textwidth]{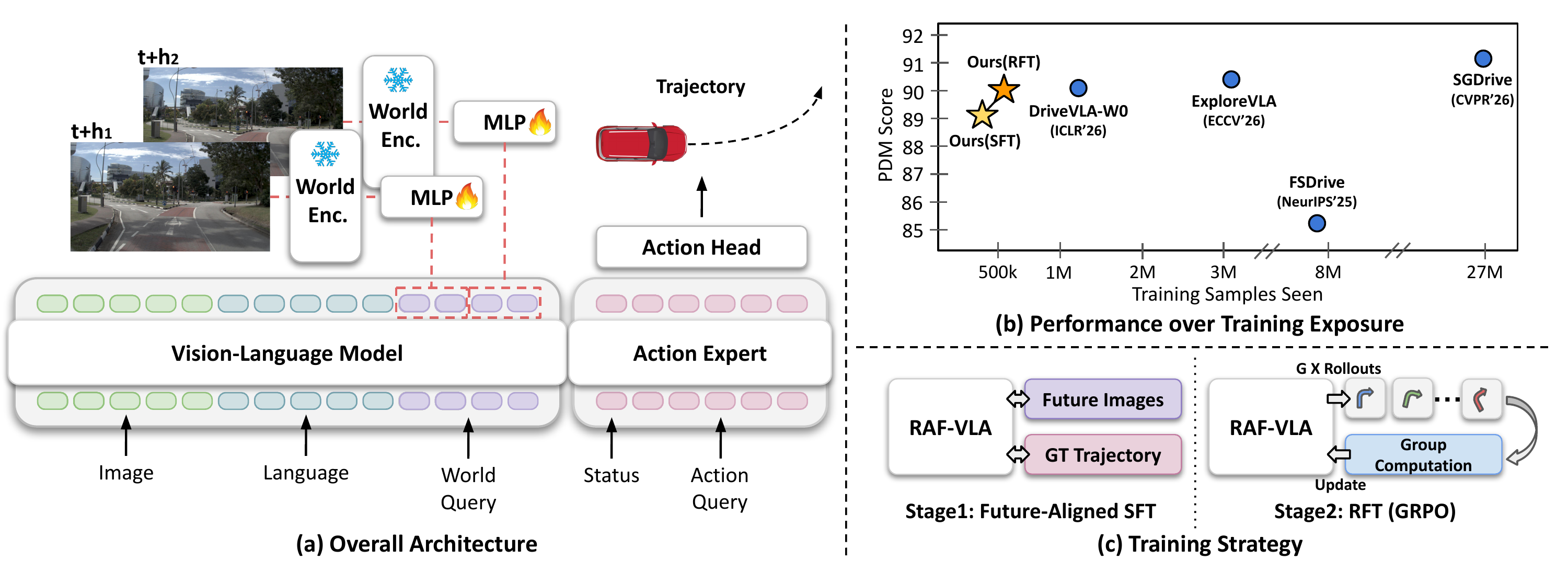}%
    \captionof{figure}{Overview of RAF-VLA. (a) Overall architecture of RAF-VLA, where future-frame representations directly guide the policy's hidden states to shape planning-relevant representations. (b) RAF-VLA achieves
    competitive planning performance with substantially fewer training samples
    seen than recent VLA planners that learn to generate future scenes.
    (c) Two-stage training strategy consisting of Future-Aligned SFT followed
    by GRPO-based RFT.}
    \label{fig:architecture}%
  \end{minipage}%
  \vspace{0.4em}%
}

\begin{document}

\maketitle
\flushbottom
\thispagestyle{empty}
\pagestyle{empty}

\begin{abstract}

Recent Vision-Language-Action (VLA) models for autonomous driving have incorporated world modeling by predicting future driving scenes alongside driving actions, demonstrating strong planning performance. Future driving scenes are utilized as dense supervision, encouraging the policy to learn rich internal representations useful for planning. However, these World-Modeling VLAs rely on explicit future generation to learn such representations, thereby introducing two key limitations: additional training burden and inference latency. To address these limitations, we propose RAF-VLA (Representation Alignment with the Future), a VLA-based autonomous driving framework that shapes planning-relevant internal representations through direct guidance from future-frame representations. RAF-VLA employs Future-Aligned Supervised Fine-Tuning, in which a straightforward regularization aligns the policy's hidden states with future-frame representations obtained from a pretrained world encoder while learning driving actions. This simple alignment allows RAF-VLA to avoid the training burden and inference latency associated with future generation. Extensive experiments on the NAVSIM benchmark show that RAF-VLA achieves competitive planning performance against state-of-the-art VLA planners with substantially fewer training samples seen. Moreover, RAF-VLA incurs only 3.8\% training overhead and a negligible 1\,ms inference overhead.

\end{abstract}

\section{INTRODUCTION}

Vision-Language-Action (VLA) models have emerged as a promising approach to end-to-end autonomous driving ~\cite{chen2024end} by fine-tuning Vision-Language Models (VLMs) pretrained on Internet-scale data as driving policies~\cite{jiang2025survey,brohan2023rt}. These models leverage VLMs' world knowledge and reasoning capabilities to map visual observations and navigation commands to driving actions. Early VLA approaches directly predicted actions from current observations~\cite{shao2024lmdrive,fu2025orion,renz2025simlingo,zhou2026opendrivevla}, whereas subsequent studies introduced textual reasoning before action prediction to harness VLMs' reasoning capabilities~\cite{hwang2024emma,luo2025adathinkdrive,zhou2026autovla,xiong2026recogdrive,yuan2026autodrive,zawalski2024robotic}. However, these approaches compress continuous visual information into discrete language, which can obscure fine-grained spatial and temporal cues important for planning. Recent VLA approaches have incorporated world modeling into policy learning, predicting future driving scenes as images, depth, or occupancy alongside action prediction~\cite{chen2025drivinggpt,zeng2026futuresightdrive,zhao2026forecasting,li2026drivevla,li2026sgdrive,sheng2026explorevla,zhao2025cot}. These World-Modeling VLAs utilize future driving scenes that reflect spatiotemporal structure and driving dynamics as dense supervision. Such future supervision encourages the policy to learn rich internal representations useful for planning, and World-Modeling VLAs have demonstrated strong planning performance.

However, explicit future generation in existing World-Modeling VLAs introduces two key limitations: additional training burden and inference latency. First, these methods jointly learn autoregressive generation, diffusion-based generation, or dense future-scene prediction alongside action prediction, which requires optimizing these additional generative objectives~\cite{zeng2026futuresightdrive,zhao2026forecasting,li2026drivevla,li2026sgdrive,sheng2026explorevla}. Second, predicting future scenes at inference requires additional steps, including sequential decoding or iterative denoising, imposing considerable latency overhead on real-time driving systems~\cite{zeng2026futuresightdrive,zhao2026forecasting,li2026sgdrive,sheng2026explorevla}. Some methods bypass future generation during inference to avoid this overhead; however, the additional training burden remains~\cite{li2026drivevla}.

To address these limitations, we reconsider the reliance on explicit future generation when leveraging future-scene supervision to learn planning-relevant representations. Prior studies on generative modeling show that while generative objectives can induce useful internal representations that transfer to downstream tasks~\cite{li2023your,tang2023emergent,chen2025deconstructing,jeong2026view,chen2020generative,dong2024dreamllm,su2026generation}, such internal representations can be shaped more directly through guidance from target representations~\cite{yu2024representation,leng2025repa,singh2025matters,xie2026unleashing}. In driving policy learning, where the ultimate goal is planning rather than future generation, these observations motivate exploiting future-scene representations to directly shape the policy's internal representations without a generation objective. To this end, we propose \textbf{RAF-VLA} (\textbf{R}epresentation \textbf{A}lignment with the \textbf{F}uture), a VLA-based autonomous driving framework that incorporates future-scene supervision at the representation level. RAF-VLA employs \textbf{Future-Aligned Supervised Fine-Tuning}, in which the policy learns driving actions while a straightforward regularization aligns its hidden states with future-frame representations. This simple alignment enables RAF-VLA to learn representations useful for planning without the training burden and inference latency of future generation.

We show that RAF-VLA achieves competitive planning performance against state-of-the-art VLA planners on the NAVSIM benchmark, with substantially fewer training samples seen than existing World-Modeling VLAs. Controlled comparisons and qualitative results further support the effectiveness of the proposed approach. Future-Aligned SFT incurs only a 3.8\% training overhead, while latency measurements show that RAF-VLA introduces a negligible 1\,ms inference overhead. The main contributions of this work are summarized as follows:

\begin{itemize}
    \item We propose RAF-VLA, a VLA-based autonomous driving framework that leverages future-scene supervision at the representation level, enabling the driving policy to directly learn planning-relevant internal representations.
    \item RAF-VLA performs Future-Aligned Supervised Fine-Tuning with a straightforward regularization on action learning that aligns the policy's hidden states with future-frame representations, thereby avoiding the training burden and inference latency of future generation.
    \item Experiments on the NAVSIM benchmark demonstrate competitive planning performance against state-of-the-art VLA planners, with substantially fewer training samples seen, low training overhead, and negligible inference overhead.

\end{itemize}

\begin{figure*}[!t]
\centering
\includegraphics[width=\textwidth]{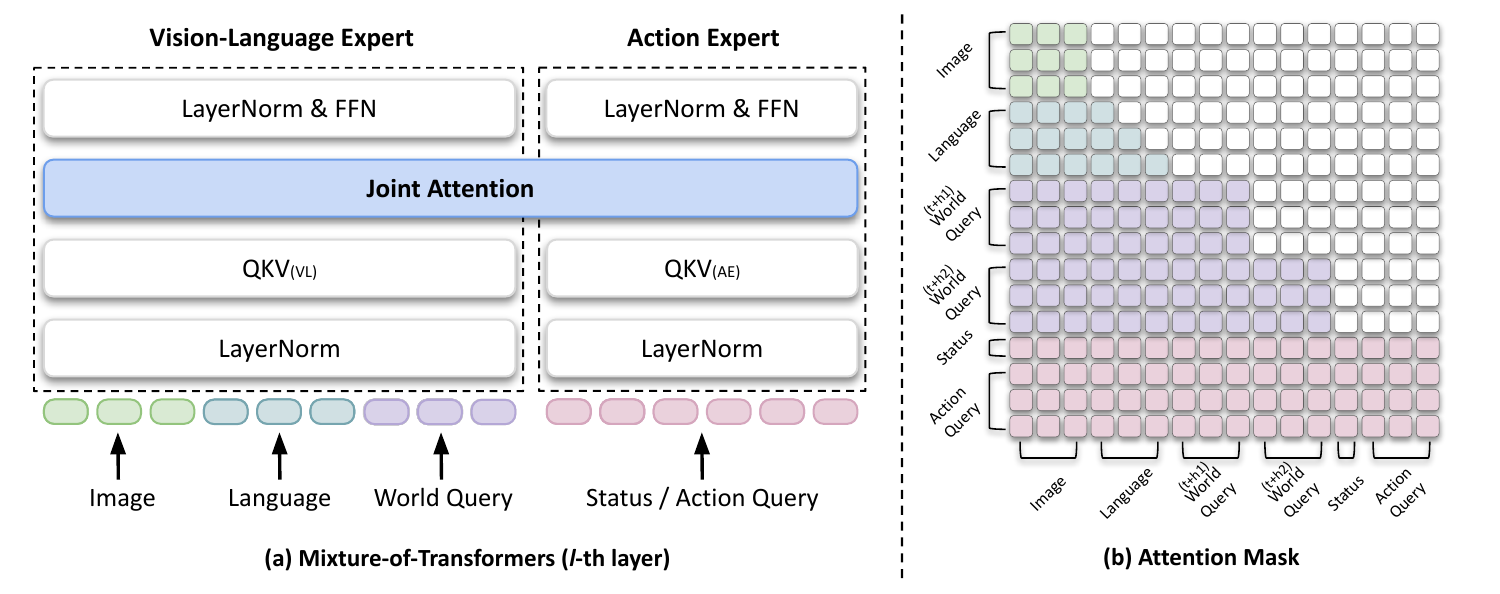}
\caption{Mixture-of-Transformers architecture. (a) Joint attention between the two experts. (b) Attention mask across token groups.}
\label{fig:attention_pattern}
\end{figure*}

\section{RELATED WORK}

\subsection{VLA for Autonomous Driving}
Vision-Language-Action (VLA) models fine-tune pretrained VLMs as action policies, leveraging their world knowledge and reasoning capabilities to improve generalization~\cite{brohan2023rt,kim2024openvla,black2024pi_0,bjorck2025gr00t}. This paradigm has also shown promise in end-to-end autonomous driving, mapping visual observations and navigation commands to driving actions to address long-tail and safety-critical scenarios~\cite{jiang2025survey}. Early approaches focused on directly predicting driving actions from observations~\cite{shao2024lmdrive,fu2025orion,renz2025simlingo,zhou2026opendrivevla}, while subsequent studies introduced textual reasoning before action prediction to further harness VLMs' reasoning capabilities, improving planning performance and generalization~\cite{hwang2024emma,luo2025adathinkdrive,zhou2026autovla,xiong2026recogdrive,yuan2026autodrive,zawalski2024robotic}. Nevertheless, these approaches rely on autoregressive language-token decoding, increasing inference latency. Generating high-quality reasoning traces also incurs additional annotation costs~\cite{rawal2026nord}. Moreover, symbolically compressing visual information into language can obscure spatiotemporal relationships and omit fine-grained visual cues important for planning~\cite{zeng2026futuresightdrive}.

Recent studies have incorporated world modeling into VLA policy learning by predicting future driving scenes alongside actions~\cite{chen2025drivinggpt,zeng2026futuresightdrive,zhao2026forecasting,li2026drivevla,li2026sgdrive,sheng2026explorevla}. These World-Modeling VLAs utilize future scenes that reflect spatiotemporal structure and driving dynamics as dense supervision to shape useful internal representations for planning and have demonstrated strong performance~\cite{li2026drivevla,li2026sgdrive,sheng2026explorevla}. DriveVLA-W0~\cite{li2026drivevla} generates future images through autoregressive visual-token prediction or a diffusion-based generation module, while ExploreVLA~\cite{sheng2026explorevla} jointly generates future RGB and depth images through masked-token prediction. SGDrive~\cite{li2026sgdrive} predicts future occupancy, agent states, and driving goals using task-specific prediction heads. However, exploiting future-scene supervision through future generation introduces additional training burden and inference overhead. This entails optimizing additional objectives for future generation, often also requiring dedicated training stages. Inference-time generation adds latency through autoregressive decoding or iterative denoising. Our work instead leverages future-scene supervision directly at the representation level, enabling the driving policy to learn planning-relevant internal representations without future generation.

\subsection{Representation Learning in Generative Models}

Prior studies have shown that generative objectives can induce useful internal representations for downstream tasks beyond generation~\cite{li2023your,tang2023emergent,jeong2026view,chen2020generative,dong2024dreamllm}. More recent studies demonstrate that such representations can be shaped through direct guidance from target representations. REPA~\cite{yu2024representation,leng2025repa} aligns internal diffusion features with representations from pretrained visual encoders to improve training efficiency, while iREPA~\cite{singh2025matters} highlights the importance of spatial structure in target representations. ARRA~\cite{xie2026unleashing} extends this guidance to autoregressive models. Whereas these methods pair representation guidance with generative objectives, our work uses representation guidance alone to shape planning-relevant representations for driving policies. Specifically, we align the VLA's hidden representations with future-frame representations alongside action prediction. This leverages future-scene supervision without future generation, thereby avoiding its associated training burden and inference overhead.

\section{METHOD}

RAF-VLA leverages future-scene supervision at the representation level to directly guide the shaping of planning-relevant representations within the driving policy. In this section, we first formulate the driving task in Sec.~\ref{sec:problem_formulation} and describe the overall architecture in Sec.~\ref{sec:overall_architecture}. We then introduce Future-Aligned Supervised Fine-Tuning (SFT) in Sec.~\ref{sec:future_aligned_sft}, which augments action learning with a straightforward regularization that aligns the policy's hidden states with future-frame representations. Finally, we describe GRPO-based reinforcement fine-tuning (RFT) in Sec.~\ref{sec:rft} to further optimize the policy using planning rewards.

\subsection{Problem Formulation}
\label{sec:problem_formulation}

\subsubsection{Input}
At time step $t$, the driving policy $\pi_{\theta}$ takes a front-view image sequence, the current navigation instruction, and the driving status as input. Let $\mathbf{I}_{t}=\{I_{t-T},\ldots,I_{t}\}$ denote the front-view image sequence including the current frame $I_{t}$, where $T$ is the observation-history length. The current navigation instruction is denoted by $\ell_{t}$. The driving status $\mathbf{s}_{t}$ consists of recent ego poses and a one-hot encoded current navigation command. We denote the driving input at time step $t$ by $q_{t}=(\mathbf{I}_{t},\ell_{t},\mathbf{s}_{t})$.

\subsubsection{Output}
Given the input $q_{t}$, the driving policy $\pi_{\theta}$ predicts an ego trajectory consisting of future waypoints: 
\begin{equation}
\hat{\boldsymbol{\tau}}_{t}
=
\pi_{\theta}(q_{t})
=
\left\{
\hat{\boldsymbol{p}}_{t+h}
\right\}_{h=1}^{H}.
\label{eq:trajectory_prediction}
\end{equation}
Here, $\hat{\boldsymbol{\tau}}_{t}$ denotes the predicted ego trajectory, $H$ is the number of future steps in the planning horizon, and $h\in\{1,\ldots,H\}$ is the planning-step index. Each predicted ego pose is defined as $\hat{\boldsymbol{p}}_{t+h}=(\hat{x}_{t+h},\hat{y}_{t+h},\hat{\theta}_{t+h})$, where $\hat{x}_{t+h}$ and $\hat{y}_{t+h}$ denote the predicted ego position at the $h$-th future step, and $\hat{\theta}_{t+h}$ denotes the corresponding heading.

\subsection{Overall Architecture}
\label{sec:overall_architecture}

As shown in Fig.~\ref{fig:architecture}(a), RAF-VLA adopts a Mixture-of-Transformers (MoT) architecture comprising a Vision-Language Expert and a lightweight Action Expert~\cite{black2024pi_0,li2026drivevla}. The two experts interact through layer-wise joint attention, enabling the Action Expert to leverage the Vision-Language Expert's multimodal context for planning.

\subsubsection{Vision-Language Expert}
The Vision-Language Expert is built on a pretrained VLM to provide multimodal driving context. Its inputs, the front-view image sequence $\mathbf{I}_t$ and navigation instruction $\ell_t$, are encoded by the VLM's vision encoder and tokenizer, respectively. We append learnable world queries after the image and language tokens as representation slots for future alignment (Sec.~\ref{sec:future_aligned_sft}). The resulting sequence is processed to form contextualized multimodal representations.

\subsubsection{Action Expert}
The Action Expert uses a similar transformer block structure to the Vision-Language Expert, with a smaller hidden dimension. Its inputs comprise a set of learnable action queries and a status token obtained by projecting the driving status $\mathbf{s}_t$ through a multilayer perceptron (MLP). The expert leverages the Vision-Language Expert's multimodal representations through joint attention to form action representations for trajectory prediction.

\subsubsection{Joint Attention}
As illustrated in Fig.~\ref{fig:attention_pattern}(a), the two experts compute their own query, key, and value matrices at each transformer layer $l$. These matrices share a common attention embedding dimension and are concatenated along the token sequence dimension to form the inputs to joint attention:
\begingroup
\thinmuskip=1mu
\thickmuskip=1mu
\begin{equation}
\mathbf{Q}^{l}=[\mathbf{Q}_{\mathrm{VL}}^{l};\mathbf{Q}_{\mathrm{AE}}^{l}],\,\mathbf{K}^{l}=[\mathbf{K}_{\mathrm{VL}}^{l};\mathbf{K}_{\mathrm{AE}}^{l}],\,\mathbf{V}^{l}=[\mathbf{V}_{\mathrm{VL}}^{l};\mathbf{V}_{\mathrm{AE}}^{l}].
\label{eq:joint_attention}
\end{equation}
\endgroup
Here, the subscripts $\mathrm{VL}$ and $\mathrm{AE}$ denote the Vision-Language Expert and Action Expert, respectively, and the semicolon denotes concatenation. As shown in Fig.~\ref{fig:attention_pattern}(b), the Action Expert applies full bidirectional attention among its status token and action queries. Attention from the Action Expert to the Vision-Language Expert is permitted, while attention in the opposite direction is masked. The resulting attention output is then split and passed back to the corresponding expert. Following prior work~\cite{li2026drivevla,li2024hydra}, we use a query-based action head with multiple trajectory anchors to decode the final action representations into an ego trajectory. The corresponding trajectory prediction loss is denoted by $\mathcal{L}_{\mathrm{act}}$~\cite{li2026drivevla}.

\subsection{Future-Aligned Supervised Fine-Tuning (SFT)}
\label{sec:future_aligned_sft}

Future-Aligned SFT augments action learning with a straightforward regularization that aligns the policy's hidden states with future-frame representations. This regularization directly guides the policy to learn planning-relevant representations without a separate generation objective.

\subsubsection{World Query}
World queries serve as representation slots for future-frame supervision. Let $\mathcal{H}_{\mathrm{F}}\subseteq\{1,\ldots,H\}$ denote the future planning steps selected for alignment. We assign $K$ learnable world queries to each $h\in\mathcal{H}_{\mathrm{F}}$. We denote their concatenated hidden states after processing by the Vision-Language Expert as $\mathbf{Z}_{t,h}\in\mathbb{R}^{Kd}$, where each hidden state is $d$-dimensional.

As shown in Fig.~\ref{fig:attention_pattern}(b), world queries can attend to the image and language tokens. We apply blockwise causal attention to the world queries, with full bidirectional attention among the $K$ queries assigned to each horizon, while masking attention to later-horizon queries.

\subsubsection{Representation Alignment with the Future}
As illustrated in Fig.~\ref{fig:architecture}(a), a frozen pretrained world encoder $f$ extracts target representations from future frames $I_{t+h}$. These representations are used solely as alignment targets during training; neither the future frames nor their extracted representations are provided as inputs to the policy. A trainable MLP projector $g_\phi$ maps the world-query hidden states $\mathbf{Z}_{t,h}$ to the target feature space. We align the projected hidden states with the corresponding future-frame representations by minimizing their mean squared error:
\begin{equation}
\mathcal{L}_{\mathrm{align}}
=
\frac{1}{|\mathcal{H}_{\mathrm{F}}|}
\sum_{h\in\mathcal{H}_{\mathrm{F}}}
\left\|
f(I_{t+h})
-
g_\phi(\mathbf{Z}_{t,h})
\right\|_2^2.
\label{eq:alignment_loss}
\end{equation}

\subsubsection{Training Objective}
We augment the trajectory prediction objective with the alignment regularization:
\begin{equation}
\mathcal{L}_{\mathrm{SFT}}
=
\mathcal{L}_{\mathrm{act}}
+
\lambda_{\mathrm{align}}\mathcal{L}_{\mathrm{align}},
\label{eq:sft_loss}
\end{equation}
where $\lambda_{\mathrm{align}}$ controls the regularization weight. The world encoder $f$ remains frozen during SFT.

The alignment regularization serves as training-time guidance to shape planning-relevant representations within the policy. The projector's outputs are used solely to compute the alignment loss. After Future-Aligned SFT, the world encoder and alignment projector are removed, while the learned world queries remain within the Vision-Language Expert. The Action Expert attends directly to their contextualized hidden states through joint attention for trajectory prediction.

\subsection{Reinforcement Fine-Tuning (RFT)}
\label{sec:rft}

After Future-Aligned SFT, we perform reinforcement fine-tuning (RFT) using Group Relative Policy Optimization (GRPO)~\cite{shao2024deepseekmath} (Fig.~\ref{fig:architecture}(c)), following the standard training strategy of recent VLA-based driving methods. GRPO improves the policy by comparing multiple candidate trajectories under the same driving context. In this stage, future-frame supervision is no longer used, and the policy is optimized with planning rewards.

Given the driving input $q_t$, the old policy $\pi_{\theta_{\mathrm{old}}}$ samples a group of $G$ candidate outputs $\mathcal{O}=\{o_1,\ldots,o_G\}$, where $o_i\sim\pi_{\theta_{\mathrm{old}}}(\cdot\mid q_t)$ denotes a sampled ego trajectory defined as in Eq. (1). The current policy $\pi_\theta$ is then optimized using the normalized group-relative advantage $A_i$ by maximizing the following objective:
\begin{equation}
\mathcal{J}(\theta)
=
\mathbb{E}_{q_t,\mathcal{O}}
\left[
\frac{1}{G}
\sum_{i=1}^{G}
\left(
\mathcal{J}_i
-
\beta D_{\mathrm{KL}}(\pi_\theta\,\|\,\pi_{\mathrm{ref}})
\right)
\right],
\label{eq:grpo_objective}
\end{equation}

\begin{equation}
\mathcal{J}_i
=
\min\left(
c_i A_i,\,
\operatorname{clip}(c_i,1-\epsilon,1+\epsilon)A_i
\right),
\label{eq:grpo_surrogate}
\end{equation}

\begin{equation}
A_i
=
\frac{
r_i-\operatorname{mean}(\{r_j\}_{j=1}^{G})
}{
\operatorname{std}(\{r_j\}_{j=1}^{G})
}.
\label{eq:grpo_advantage}
\end{equation}

Here, $\mathcal{J}_i$ denotes the clipped surrogate objective, and $c_i=\frac{\pi_\theta(o_i\mid q_t)}{\pi_{\theta_{\mathrm{old}}}(o_i\mid q_t)}$ is the probability ratio between the current and old policies. The group-relative advantage $A_i$ is computed by normalizing the planning reward $r_i$ using the mean and standard deviation of rewards within the sampled group. The hyperparameter $\epsilon$ controls clipping, while $\beta$ weights the KL penalty between the current policy and the SFT reference policy $\pi_{\mathrm{ref}}$.

The planning reward $r_i$ is computed using the NAVSIM Predictive Driver Model Score (PDMS)~\cite{dauner2024navsim}, which evaluates trajectory quality across multiple driving criteria, including collision avoidance, drivable-area compliance, progress, time-to-collision, and comfort.

\section{EXPERIMENTS}

\begin{table*}[t]
\centering
\caption{Comparison on NAVSIM v1 with closed-loop metrics. Abbreviations: Sensors: 1x Cam: front-view camera only; \\Nx Cam: N surround-view cameras; L: LiDAR. $^\dagger$: best-of-$N$ ($N=6$) strategy following~\cite{zhou2026autovla,li2026drivevla,sheng2026explorevla}.}
\label{tab:navsim_main}
\setlength{\tabcolsep}{5.55pt}
\renewcommand{\arraystretch}{0.99}
\resizebox{0.86\textwidth}{!}{%
\begin{tabular}{l|c|c|ccccc|>{\columncolor{gray!15}}c}
\toprule
Method & Venue & Sensors & NC$\uparrow$ & DAC$\uparrow$ & EP$\uparrow$ & TTC$\uparrow$ & C.$\uparrow$ & PDMS$\uparrow$ \\
\midrule
UniAD~\cite{hu2023planning} & CVPR'23 & 6x Cam & 97.8 & 91.9 & 78.8 & 92.9 & \textbf{100.0} & 83.4 \\
TransFuser~\cite{chitta2022transfuser} & TPAMI'23 & 3x Cam + L & 97.7 & 92.8 & 79.2 & 92.8 & \textbf{100.0} & 84.0 \\
LAW~\cite{li2025enhancing} & ICLR'25 & 1x Cam & 96.4 & 95.4 & 81.7 & 88.7 & 99.9 & 84.6 \\
Epona~\cite{zhang2025epona} & ICCV'25 & 1x Cam & 97.9 & 95.1 & 80.4 & 93.8 & 99.9 & 86.2 \\   
Hydra-MDP~\cite{li2024hydra} & arXiv'24 & 3x Cam + L & 98.3 & 96.0 & 78.7 & 94.6 & \textbf{100.0} & 86.5 \\
DiffusionDrive~\cite{liao2025diffusiondrive} & CVPR'25 & 3x Cam + L & 98.2 & 96.2 & 82.2 & 94.7 & \textbf{100.0} & 88.1 \\
WoTE~\cite{li2025end} & ICCV'25 & 3x Cam + L & 98.5 & 96.8 & 81.9 & 94.9 & 99.9 & 88.3 \\
DriveLaW~\cite{xia2026drivelaw} & CVPR'26 & 1x Cam & 99.0 & 97.1 & 81.3 & 96.7 & \textbf{100.0} & 89.1 \\
\midrule
\multicolumn{9}{l}{\emph{Textual-Reasoning VLA}} \\
AutoVLA~\cite{zhou2026autovla} & NeurIPS'25 & 3x Cam & 98.4 & 95.6 & 81.9 & 98.0 & 99.9 & 89.1 \\
ReCogDrive~\cite{xiong2026recogdrive} & ICLR'26 & 3x Cam & 98.2 & 97.8 & 83.5 & 95.2 & 99.8 & 89.6 \\
AdaThinkDrive~\cite{luo2025adathinkdrive} & ICRA'26 & 1x Cam & 98.4 & 97.8 & 84.4 & 95.2 & \textbf{100.0} & 90.3 \\
AutoVLA$^\dagger$~\cite{zhou2026autovla} & NeurIPS'25 & 3x Cam & 99.1 & 97.1 & 87.6 & 97.1 & \textbf{100.0} & 92.1 \\
AdaThinkDrive$^\dagger$~\cite{luo2025adathinkdrive} & ICRA'26 & 1x Cam & 99.1 & 98.8 & 87.9 & 97.2 & \textbf{100.0} & 93.0 \\
\midrule
\multicolumn{9}{l}{\emph{World-Modeling VLA}} \\
DrivingGPT~\cite{chen2025drivinggpt} & ICCV'25 & 1x Cam & 98.9 & 90.7 & 79.7 & 94.9 & 95.6 & 82.4 \\
FSDrive~\cite{zeng2026futuresightdrive} & NeurIPS'25 & 3x Cam & 98.2 & 93.8 & 80.1 & 93.3 & 99.9 & 85.1 \\
PWM~\cite{zhao2026forecasting} & NeurIPS'25 & 1x Cam & 98.6 & 95.9 & 81.8 & 95.4 & \textbf{100.0} & 88.1 \\
DriveVLA-W0~\cite{li2026drivevla} & ICLR'26 & 1x Cam & 98.7 & 99.1 & 83.3 & 95.3 & 99.3 & 90.2 \\
ExploreVLA~\cite{sheng2026explorevla} & ECCV'26 & 1x Cam & 98.8 & 98.4 & 83.5 & 96.5 & 99.9 & 90.4 \\
SGDrive~\cite{li2026sgdrive} & CVPR'26 & 1x Cam & 98.6 & 97.8 & 85.8 & 96.2 & \textbf{100.0} & 91.1 \\
DriveVLA-W0$^\dagger$~\cite{li2026drivevla} & ICLR'26 & 1x Cam & 99.3 & 97.4 & 88.3 & 97.0 & 99.9 & 93.0 \\
ExploreVLA$^\dagger$~\cite{sheng2026explorevla} & ECCV'26 & 1x Cam & \textbf{99.4} & 98.9 & 88.3 & \textbf{98.3} & 99.7 & 93.7 \\
\midrule
\rowcolor{gray!15}
\textbf{RAF-VLA (Ours)} & -- & 1x Cam & 98.0 & 98.5 & 85.2 & 93.9 & 99.7 & 90.0 \\
\rowcolor{gray!15}
\textbf{RAF-VLA$^\dagger$ (Ours)} & -- & 1x Cam & 99.0 & \textbf{99.6} & \textbf{89.8} & 97.3 & \textbf{100.0} & \textbf{94.2} \\
\bottomrule
\end{tabular}
}
\end{table*}

\begin{table*}[t]
\centering
\caption{NAVSIM v2 results with extended closed-loop metrics using Future-Aligned SFT.}
\label{tab:navsim_extended}
\setlength{\tabcolsep}{5.1pt}
\renewcommand{\arraystretch}{0.99}
\resizebox{0.86\textwidth}{!}{%
\begin{tabular}{l|ccccccccc|>{\columncolor{gray!15}}c}
\toprule
Method & NC$\uparrow$ & DAC$\uparrow$ & DDC$\uparrow$ & TLC$\uparrow$ & EP$\uparrow$ & TTC$\uparrow$ & LK$\uparrow$ & HC$\uparrow$ & EC$\uparrow$ & EPDMS$\uparrow$ \\
\midrule
TransFuser~\cite{chitta2022transfuser} & 96.9 & 89.9 & 97.8 & 99.7 & 87.1 & 95.4 & 92.7 & 98.3 & 87.2 & 76.7 \\
VADv2~\cite{chen2024vadv2} & 97.3 & 91.7 & 98.2 & \textbf{99.9} & 77.6 & 92.7 & 66.0 & \textbf{100.0} & \textbf{97.4} & 76.6 \\
Hydra-MDP~\cite{li2024hydra} & 97.5 & 96.3 & 98.3 & \textbf{99.9} & 80.1 & 93.0 & 65.5 & \textbf{100.0} & \textbf{97.4} & 79.8 \\
Hydra-MDP++~\cite{li2025hydra} & 97.2 & 97.5 & 99.4 & 99.6 & 83.1 & 96.5 & 94.4 & 98.2 & 70.9 & 81.4 \\
ReCogDrive~\cite{xiong2026recogdrive} & 98.3 & 95.2 & \textbf{99.5} & 99.8 & 87.1 & 97.5 & 96.6 & 98.3 & 86.5 & 83.6 \\
DiffusionDrive~\cite{liao2025diffusiondrive} & 98.2 & 95.9 & 99.4 & 99.8 & \textbf{87.5} & 97.3 & \textbf{96.8} & 98.3 & 87.7 & 84.5 \\
DriveVLA-W0~\cite{li2026drivevla} & 98.5 & \textbf{99.1} & 98.0 & 99.7 & 86.4 & \textbf{98.1} & 93.2 & 97.9 & 58.9 & 86.1 \\
SGDrive~\cite{li2026sgdrive} & \textbf{98.6} & 94.3 & \textbf{99.5} & \textbf{99.9} & 86.0 & 97.9 & 96.1 & 98.3 & 85.9 & \textbf{86.2} \\
\midrule
\rowcolor{gray!15}
\textbf{RAF-VLA (Ours)} & 97.5 & 98.2 & 99.4 & 99.8 & 86.9 & 97.0 & 92.7 & 98.0 & 65.7 & \textbf{86.2} \\
\bottomrule
\end{tabular}
}
\end{table*}

\begin{figure*}[t!]
\centering
\includegraphics[width=\textwidth]{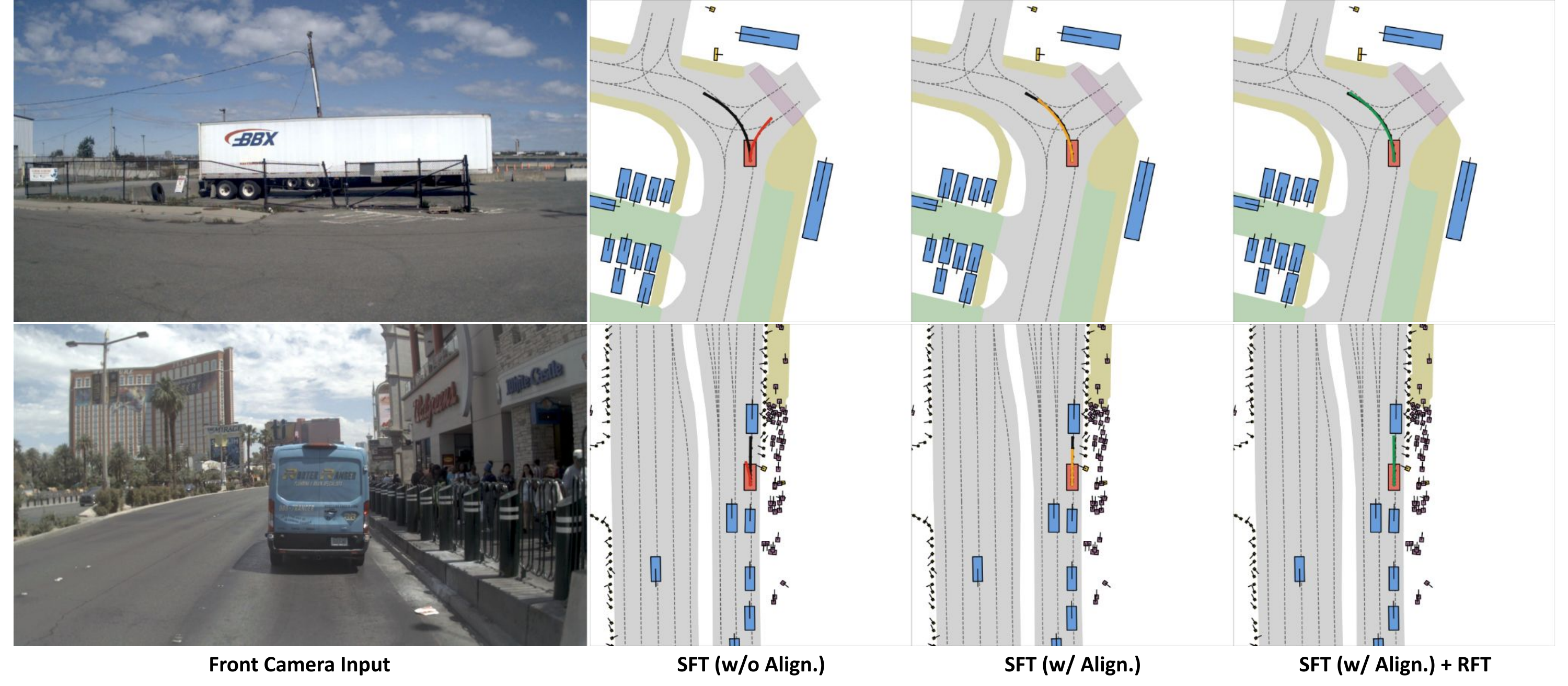}
\caption{Qualitative comparison of trajectory predictions on the NAVSIM \texttt{navtest} split. From left to right: front-camera input, vanilla SFT without alignment, Future-Aligned SFT, and Future-Aligned SFT followed by GRPO-based RFT (black: GT, red/orange/green: prediction from each model).}
\label{fig:qualitative_results}
\end{figure*}
\begin{table}[t]
\centering
\caption{Ablation of future-representation alignment under SFT and RFT. Bold values indicate the best result for each metric.}
\label{tab:alignment_rft_factorial}
\fontsize{8}{9}\selectfont
\setlength{\tabcolsep}{1.8pt}
\begin{lrbox}{\rafalignmenttable}
\begin{tabular}{c|ccc|ccccc|>{\columncolor{gray!15}}c}
\toprule
ID & Alignment & SFT & RFT & NC$\uparrow$ & DAC$\uparrow$ & EP$\uparrow$ & TTC$\uparrow$ & C.$\uparrow$ & PDMS$\uparrow$ \\
\midrule
1 &  & $\checkmark$ &  & 97.86 & 98.58 & 81.01 & 93.03 & 99.52 & 87.92 \\
2 &  & $\checkmark$ & $\checkmark$ & 97.42 & 98.39 & 84.31 & 92.38 & 99.51 & 88.85 \\
\midrule
3 & $\checkmark$ & $\checkmark$ &  & \textbf{98.04} & \textbf{98.71} & 82.99 & 93.60 & \textbf{99.79} & 89.13 \\
4 & $\checkmark$ & $\checkmark$ & $\checkmark$ & 97.98 & 98.46 & \textbf{85.21} & \textbf{93.86} & 99.73 & \textbf{90.03} \\
\bottomrule
\end{tabular}
\end{lrbox}
\global\rafablationwidth=\wd\rafalignmenttable
\usebox{\rafalignmenttable}
\end{table}

\begin{table}[t]
\centering
\caption{Inference latency with and without alignment. Inference time denotes the average time required to predict 4-second trajectories on the NAVSIM v1 \texttt{navtest} split.}
\label{tab:inference_latency}
\setlength{\tabcolsep}{10pt}
\fontsize{8}{9}\selectfont
\begin{tabular}{c|c|c}
\toprule
Alignment & Latency (s)$\downarrow$ & $\Delta$ \\
\midrule
$\times$ & 0.152 & -- \\
$\checkmark$ & 0.153 & \textbf{+0.001} \\
\bottomrule
\end{tabular}
\end{table}

\newcommand{\rafablationtables}{%
\begin{table}[t]
\begin{minipage}[t]{\columnwidth}
\centering
\captionof{table}{Ablation of future-alignment horizons on NAVSIM v1.}
\label{tab:horizon_ablation}
\fontsize{8}{9}\selectfont
\setlength{\tabcolsep}{1.6pt}
\begin{tabular*}{\rafablationwidth}{@{\extracolsep{\fill}}c|*{4}{>{\centering\arraybackslash}p{11pt}}@{\hspace{3pt}}|ccccc!{\hspace{1pt}\vrule width\arrayrulewidth\hspace{-1pt}}>{\columncolor{gray!15}}c}
\toprule
ID & 1\,s & 2\,s & 3\,s & 4\,s & NC$\uparrow$ & DAC$\uparrow$ & EP$\uparrow$ & TTC$\uparrow$ & C.$\uparrow$ & PDMS$\uparrow$ \\
\midrule
1 &  &  &  &  & 97.86 & 98.58 & 81.01 & 93.03 & 99.52 & 87.92 \\
2 & $\checkmark$ &  &  &  & \textbf{98.40} & 98.50 & 82.33 & \textbf{93.91} & \textbf{99.81} & 88.93 \\
3 &  &  &  & $\checkmark$ & 98.30 & 98.67 & 82.39 & 93.78 & 99.76 & 88.97 \\
4 & $\checkmark$ & $\checkmark$ &  &  & 97.80 & 98.48 & 82.32 & 93.47 & 99.53 & 88.60 \\
5 & $\checkmark$ &  &  & $\checkmark$ & 98.04 & \textbf{98.71} & \textbf{82.99} & 93.60 & 99.79 & \textbf{89.13} \\
6 & $\checkmark$ & $\checkmark$ & $\checkmark$ & $\checkmark$ & 97.88 & 98.53 & 82.52 & 93.13 & 99.79 & 88.62 \\
\bottomrule
\end{tabular*}
\end{minipage}
\par\vspace{\floatsep}
\begin{minipage}[t]{\columnwidth}
\centering
\captionof{table}{Ablation of the query count $K$ on NAVSIM v1.}
\label{tab:k_ablation}
\fontsize{8}{9}\selectfont
\setlength{\tabcolsep}{1pt}
\begin{tabularx}{\rafablationwidth}{c|>{\centering\arraybackslash}p{18pt}|*{5}{>{\centering\arraybackslash}X}|>{\columncolor{gray!15}\centering\arraybackslash}X}
\toprule
ID & $K$ & NC$\uparrow$ & DAC$\uparrow$ & EP$\uparrow$ & TTC$\uparrow$ & C.$\uparrow$ & PDMS$\uparrow$ \\
\midrule
1 & 2  & \textbf{98.14} & 98.36 & 82.94 & 93.21 & 99.78 & 88.76 \\
2 & 4  & 98.04 & \textbf{98.71} & \textbf{82.99} & \textbf{93.60} & \textbf{99.79} & \textbf{89.13} \\
3 & 8  & 97.94 & 98.63 & 82.30 & 93.25 & 99.76 & 88.64 \\
4 & 16 & 97.97 & 98.63 & 82.49 & 93.03 & \textbf{99.79} & 88.65 \\
\bottomrule
\end{tabularx}
\end{minipage}
\end{table}
}

\subsection{Dataset and Metrics}

\subsubsection{Dataset}
We conduct extensive experiments on the NAVSIM v1 and NAVSIM v2 benchmarks~\cite{dauner2024navsim,cao2025pseudo} to evaluate 
the performance of end-to-end autonomous driving models. NAVSIM is built on OpenScene~\cite{openscene2023}, which provides real-world sensor observations and annotated driving scenes. The benchmark filters out near-trivial driving cases to emphasize challenging scenarios with a higher proportion of non-trivial maneuvers. We train RAF-VLA on the \emph{navtrain} split and evaluate on the \emph{navtest} split.

\subsubsection{Metrics}
For NAVSIM v1, we evaluate planning performance using the Predictive Driver Model Score (PDMS):
\begin{equation}
\mathrm{PDMS}
=
\mathrm{NC}\times\mathrm{DAC}\times
\frac{5\mathrm{EP}+5\mathrm{TTC}+2\mathrm{C}}{12},
\label{eq:pdms}
\end{equation}
where No At-Fault Collision (NC), Drivable Area Compliance (DAC), Ego Progress (EP), Time-to-Collision (TTC), and Comfort (C) constitute the evaluation metrics.

For NAVSIM v2, we use the Extended Predictive Driver Model Score (EPDMS):
\begin{equation}
\begin{aligned}
\mathrm{EPDMS} &= \mathrm{NC}\times\mathrm{DAC}\times\mathrm{DDC}\times\mathrm{TLC}\\
&\times \frac{5\mathrm{EP}+5\mathrm{TTC}+2\mathrm{LK}+2\mathrm{HC}+2\mathrm{EC}}{16},
\end{aligned}
\label{eq:epdms}
\end{equation}
which additionally includes Driving Direction Compliance (DDC), Traffic Light Compliance (TLC), Lane Keeping (LK), History Comfort (HC), and Extended Comfort (EC). Higher values indicate better performance for all metrics.

\subsection{Implementation Details}
\label{sec:implementation_details}

We use Qwen2.5-VL-7B-Instruct as the Vision-Language Expert~\cite{bai2025qwen25vltechnicalreport}, a lightweight transformer with an embedding dimension of 1024 as the Action Expert, and a pretrained encoder from the Cosmos world model~\cite{agarwal2025cosmos} to extract target representations from future frames. The alignment projector is a two-layer MLP with a GELU activation~\cite{hendrycks2016gaussian}.
For Future-Aligned SFT, we train for 4 epochs with AdamW, a global batch size of 144, and a learning rate of $1\times10^{-5}$ with cosine decay, while freezing the VLM vision encoder and world encoder. Unless otherwise specified, alignment uses the 1\,s and 4\,s horizons with $K=4$ world queries per horizon and $\lambda_{\mathrm{align}}=1$.
For RFT, we perform GRPO for 1.5 epochs with a global batch size of 288, a learning rate of $2\times10^{-6}$, group size $G=8$, and KL coefficient $\beta=0.04$; the world encoder and alignment projector are removed, while the VLM vision encoder, world queries, and action head remain frozen. Both stages are trained on 4$\times$H200 GPUs, taking approximately 11 hours for SFT and 7 hours for RFT.

\subsection{Main Results}

\subsubsection{Performance Comparison}
Table~\ref{tab:navsim_main} compares RAF-VLA with state-of-the-art (SOTA) methods on NAVSIM v1. Using only a single front-view camera, RAF-VLA achieves a PDMS of 90.0, demonstrating competitive planning performance against SOTA VLA-based planners. As shown in Fig.~\ref{fig:architecture}(b), this performance is achieved with substantially fewer training samples seen than existing World-Modeling VLAs. The total training samples seen for each method are estimated from its reported configurations, official implementations, and official dataset splits. With the best-of-$N$ strategy, the PDMS further increases to 94.2, achieving the highest overall performance, together with the best DAC and EP scores of 99.6 and 89.8, respectively. We further evaluate the Future-Aligned SFT model on NAVSIM v2 using the extended metric. As shown in Table~\ref{tab:navsim_extended}, it achieves an EPDMS of 86.2, matching the best reported SFT result.

\subsubsection{Effectiveness of Future-Aligned SFT}
Table~\ref{tab:alignment_rft_factorial} evaluates the effect of the regularization for representation alignment in Future-Aligned SFT, both before and after reinforcement fine-tuning. For the vanilla SFT baseline, we remove the entire future-alignment mechanism, including the associated world queries. Future-Aligned SFT improves the PDMS from 87.92 to 89.13 (+1.21) over SFT without alignment. This improvement incurs only a 3.8\% increase in training time, from 10.56 h (42.2 H200 GPU-hours) for vanilla SFT to 10.96 h (43.8 H200 GPU-hours) for Future-Aligned SFT, including online extraction of future representations. Importantly, this gain is retained after reinforcement fine-tuning. The model initialized from Future-Aligned SFT achieves a PDMS of 90.03, compared with 88.85 for the model initialized from SFT without alignment (+1.18). These results show that the benefit introduced during Future-Aligned SFT persists through subsequent RFT.

Fig.~\ref{fig:qualitative_results} provides a qualitative comparison of trajectory predictions from SFT without alignment, Future-Aligned SFT, and Future-Aligned SFT followed by RFT. SFT without alignment produces trajectories that poorly follow the lane consistent with the navigation command and deviate from the ground-truth trajectory. They also fail to make sufficient progress toward the intended direction. Future-Aligned SFT corrects these failures, yielding trajectories that better follow the intended maneuver, while subsequent RFT further refines the predicted paths. The quantitative and qualitative results consistently support the effectiveness of Future-Aligned SFT.

\subsubsection{Inference Overhead}
As shown in Table~\ref{tab:inference_latency}, Future-Aligned SFT adds only 0.001\,s (1\,ms) of latency for predicting a 4-second trajectory, resulting in negligible inference overhead. Inference latency is measured on a single NVIDIA H200 GPU with a batch size of 1. This is because Future-Aligned SFT directly shapes policy representations through alignment during training, leaving only a few learned world queries to be processed during the prefill stage at inference. No additional scene prediction or generation is required.

\rafablationtables
\subsection{Ablation Studies}

Unless otherwise specified, all ablation studies follow the implementation details in Sec.~\ref{sec:implementation_details} and are conducted under the SFT setting.

\subsubsection{Effect of Future Alignment Horizons}
Table~\ref{tab:horizon_ablation} compares different future horizons used for representation alignment. Using either the 1\,s or 4\,s future representation improves the planning performance over the vanilla SFT baseline, with the 4\,s horizon corresponding to the end of the planning horizon. Among the evaluated configurations, jointly using the 1\,s and 4\,s horizons achieves the highest PDMS of 89.13, while adding intermediate horizons does not provide further improvement. We therefore use the 1\,s and 4\,s horizons as the default configuration.

\subsubsection{Effect of the Number of World Queries}
Table~\ref{tab:k_ablation} ablates the number of world queries assigned to each future horizon. Among the evaluated settings, $K=4$ achieves the highest PDMS of 89.13, while increasing the number of world queries beyond four does not provide further improvement. We therefore use four world queries per horizon as the default configuration.

\FloatBarrier

\section{CONCLUSIONS}
In this work, we introduced RAF-VLA, a VLA-based autonomous driving framework that leverages future-scene supervision at the representation level without future generation. RAF-VLA performs Future-Aligned Supervised Fine-Tuning, which aligns the policy's hidden states with future-frame representations, thereby providing direct guidance for shaping planning-relevant internal representations during action learning. Experiments on NAVSIM v1 and v2 demonstrate competitive planning performance with substantially fewer training samples seen, low training overhead, and negligible inference overhead. Controlled comparisons and qualitative results provide further evidence for the effectiveness of our proposed approach. Future work will explore alternative target encoders and projection designs. We also plan to extend future representation alignment to multi-view inputs and richer scene modalities.





\bibliographystyle{IEEEtran}
\bibliography{ref}

@article{chen2024end,
  title={End-to-end autonomous driving: Challenges and frontiers},
  author={Chen, Li and Wu, Penghao and Chitta, Kashyap and Jaeger, Bernhard and Geiger, Andreas and Li, Hongyang},
  journal={IEEE Transactions on Pattern Analysis and Machine Intelligence},
  volume={46},
  number={12},
  pages={10164--10183},
  year={2024},
  publisher={IEEE}
}

@inproceedings{jiang2025survey,
  title     = {A Survey on Vision-Language-Action Models for Autonomous Driving},
  author    = {Jiang, Sicong and Huang, Zilin and Qian, Kangan and Luo, Ziang and Zhu, Tianze and Zhong, Yang and Tang, Yihong and Kong, Menglin and Wang, Yunlong and Jiao, Siwen and others},
  booktitle = {Proceedings of the IEEE/CVF International Conference on Computer Vision},
  pages     = {4583--4595},
  year      = {2025}
}

@article{brohan2023rt,
  title={Rt-2: Vision-language-action models transfer web knowledge to robotic control},
  author={Brohan, Anthony and Brown, Noah and Carbajal, Justice and Chebotar, Yevgen and Chen, Xi and Choromanski, Krzysztof and Ding, Tianli and Driess, Danny and Dubey, Avinava and Finn, Chelsea and others},
  journal={arXiv preprint arXiv:2307.15818},
  year={2023}
}

@inproceedings{shao2024lmdrive,
  title={Lmdrive: Closed-loop end-to-end driving with large language models},
  author={Shao, Hao and Hu, Yuxuan and Wang, Letian and Song, Guanglu and Waslander, Steven L and Liu, Yu and Li, Hongsheng},
  booktitle={2024 IEEE/CVF Conference on Computer Vision and Pattern Recognition (CVPR)},
  pages={15120--15130},
  year={2024},
  organization={IEEE}
}

@inproceedings{fu2025orion,
  title={Orion: A holistic end-to-end autonomous driving framework by vision-language instructed action generation},
  author={Fu, Haoyu and Zhang, Diankun and Zhao, Zongchuang and Cui, Jianfeng and Liang, Dingkang and Zhang, Chong and Zhang, Dingyuan and Xie, Hongwei and Wang, Bing and Bai, Xiang},
  booktitle={2025 IEEE/CVF International Conference on Computer Vision (ICCV)},
  pages={24823--24834},
  year={2025},
  organization={IEEE}
}

@inproceedings{renz2025simlingo,
  title={Simlingo: Vision-only closed-loop autonomous driving with language-action alignment},
  author={Renz, Katrin and Chen, Long and Arani, Elahe and Sinavski, Oleg},
  booktitle={2025 IEEE/CVF Conference on Computer Vision and Pattern Recognition (CVPR)},
  pages={11993--12003},
  year={2025},
  organization={IEEE}
}

@inproceedings{zhou2026opendrivevla,
  title={Opendrivevla: Towards end-to-end autonomous driving with large vision language action model},
  author={Zhou, Xingcheng and Han, Xuyuan and Yang, Feng and Ma, Yunpu and Tresp, Volker and Knoll, Alois},
  booktitle={Proceedings of the AAAI Conference on Artificial Intelligence},
  volume={40},
  number={16},
  pages={13782--13790},
  year={2026}
}

@article{hwang2024emma,
  title={Emma: End-to-end multimodal model for autonomous driving},
  author={Hwang, Jyh-Jing and Xu, Runsheng and Lin, Hubert and Hung, Wei-Chih and Ji, Jingwei and Choi, Kristy and Huang, Di and He, Tong and Covington, Paul and Sapp, Benjamin and others},
  journal={arXiv preprint arXiv:2410.23262},
  year={2024}
}

@article{luo2025adathinkdrive,
  title={Adathinkdrive: Adaptive thinking via reinforcement learning for autonomous driving},
  author={Luo, Yuechen and Li, Fang and Xu, Shaoqing and Lai, Zhiyi and Yang, Lei and Chen, Qimao and Luo, Ziang and Xie, Zixun and Jiang, Shengyin and Liu, Jiaxin and others},
  journal={arXiv preprint arXiv:2509.13769},
  year={2025}
}

@article{zhou2026autovla,
  title={Autovla: A vision-language-action model for end-to-end autonomous driving with adaptive reasoning and reinforcement fine-tuning},
  author={Zhou, Zewei and Cai, Tianhui and Zhao, Seth and Zhang, Yun and Huang, Zhiyu and Zhou, Bolei and Ma, Jiaqi},
  journal={Advances in Neural Information Processing Systems},
  volume={38},
  pages={27920--27956},
  year={2026}
}

@inproceedings{xiong2026recogdrive,
  title={Recogdrive: A reinforced cognitive framework for end-to-end autonomous driving},
  author={Xiong, Kaixin and Guo, Xiangyu and Li, Fang and Yan, Sixu and Xu, Gangwei and Zhou, Lijun and Chen, Long and Sun, Haiyang and Wang, Bing and Ma, Kun and others},
  booktitle={International Conference on Learning Representations},
  volume={2026},
  pages={157518--157556},
  year={2026}
}

@inproceedings{yuan2026autodrive,
  title={Autodrive-r$^2$: Incentivizing reasoning and self-reflection capacity for VLA model in autonomous driving},
  author={Yuan, Zhenlong and Qian, Chengxuan and Tang, Jing and Chen, Rui and Song, Zijian and Sun, Lei and Chu, Xiangxiang and Cai, Yujun and Zhang, Dapeng and Li, Shuo},
  booktitle={International Conference on Learning Representations},
  volume={2026},
  pages={97934--97955},
  year={2026}
}

@article{zawalski2024robotic,
  title={Robotic control via embodied chain-of-thought reasoning},
  author={Zawalski, Micha{\l} and Chen, William and Pertsch, Karl and Mees, Oier and Finn, Chelsea and Levine, Sergey},
  journal={arXiv preprint arXiv:2407.08693},
  year={2024}
}

@inproceedings{chen2025drivinggpt,
  title={Drivinggpt: Unifying driving world modeling and planning with multi-modal autoregressive transformers},
  author={Chen, Yuntao and Wang, Yuqi and Zhang, Zhaoxiang},
  booktitle={2025 IEEE/CVF International Conference on Computer Vision (ICCV)},
  pages={26890--26900},
  year={2025},
  organization={IEEE}
}

@article{zeng2026futuresightdrive,
  title={Futuresightdrive: Thinking visually with spatio-temporal cot for autonomous driving},
  author={Zeng, Shuang and Chang, Xinyuan and Xie, Mengwei and Liu, Xinran and Bai, Yifan and Pan, Zheng and Xu, Mu and Wei, Xing},
  journal={Advances in Neural Information Processing Systems},
  volume={38},
  pages={67299--67318},
  year={2026}
}

@article{zhao2026forecasting,
  title={From forecasting to planning: Policy world model for collaborative state-action prediction},
  author={Zhao, Zhida and Fu, Talas and Wang, Yifan and Wang, Lijun and Lu, Huchuan},
  journal={Advances in Neural Information Processing Systems},
  volume={38},
  pages={134585--134611},
  year={2026}
}

@inproceedings{li2026drivevla,
  title={Drivevla-w0: World models amplify data scaling law in autonomous driving},
  author={Li, Yingyan and Shang, Shuyao and Liu, Weisong and Zhan, Bing and Wang, Haochen and Wang, Yuqi and Chen, Yuntao and Wang, Xiaoman and An, Yasong and Tang, Chufeng and others},
  booktitle={International Conference on Learning Representations},
  volume={2026},
  pages={7890--7911},
  year={2026}
}

@article{li2026sgdrive,
  title={Sgdrive: Scene-to-goal hierarchical world cognition for autonomous driving},
  author={Li, Jingyu and Wu, Junjie and Hu, Dongnan and Huang, Xiangkai and Sun, Bin and Hao, Zhihui and Lang, Xianpeng and Zhu, Xiatian and Zhang, Li},
  journal={arXiv preprint arXiv:2601.05640},
  year={2026}
}

@article{sheng2026explorevla,
  title={Explorevla: Dense world modeling and exploration for end-to-end autonomous driving},
  author={Sheng, Zihao and Ye, Xin and Luo, Jingru and Chen, Sikai and Ren, Liu},
  journal={arXiv preprint arXiv:2604.02714},
  year={2026}
}

@inproceedings{zhao2025cot,
  title={Cot-vla: Visual chain-of-thought reasoning for vision-language-action models},
  author={Zhao, Qingqing and Lu, Yao and Kim, Moo Jin and Fu, Zipeng and Zhang, Zhuoyang and Wu, Yecheng and Li, Zhaoshuo and Ma, Qianli and Han, Song and Finn, Chelsea and others},
  booktitle={2025 IEEE/CVF Conference on Computer Vision and Pattern Recognition (CVPR)},
  pages={1702--1713},
  year={2025},
  organization={IEEE}
}

@inproceedings{li2023your,
  title={Your diffusion model is secretly a zero-shot classifier},
  author={Li, Alexander C and Prabhudesai, Mihir and Duggal, Shivam and Brown, Ellis and Pathak, Deepak},
  booktitle={2023 IEEE/CVF International Conference on Computer Vision (ICCV)},
  pages={2206--2217},
  year={2023},
  organization={IEEE}
}

@article{tang2023emergent,
  title={Emergent correspondence from image diffusion},
  author={Tang, Luming and Jia, Menglin and Wang, Qianqian and Phoo, Cheng Perng and Hariharan, Bharath},
  journal={Advances in neural information processing systems},
  volume={36},
  pages={1363--1389},
  year={2023}
}

@inproceedings{chen2025deconstructing,
  title={Deconstructing denoising diffusion models for self-supervised learning},
  author={Chen, Xinlei and Liu, Zhuang and Xie, Saining and He, Kaiming},
  booktitle={International Conference on Learning Representations},
  volume={2025},
  pages={55458--55472},
  year={2025}
}

@article{jeong2026view,
  title={To View Transform or Not to View Transform: NeRF-based Pre-training Perspective},
  author={Jeong, Hyeonjun and Shin, Juyeb and Kum, Dongsuk},
  journal={arXiv preprint arXiv:2603.28090},
  year={2026}
}

@inproceedings{chen2020generative,
  title={Generative pretraining from pixels},
  author={Chen, Mark and Radford, Alec and Child, Rewon and Wu, Jeffrey and Jun, Heewoo and Luan, David and Sutskever, Ilya},
  booktitle={International conference on machine learning},
  pages={1691--1703},
  year={2020},
  organization={PMLR}
}

@inproceedings{dong2024dreamllm,
  title={Dreamllm: Synergistic multimodal comprehension and creation},
  author={Dong, Runpei and Peng, Yuang and Qi, Zekun and Ge, Zheng and Yang, Jinrong and Zhao, Liang and Sun, Jianjian and Zhou, Hongyu and Wei, Haoran and Kong, Xiangwen and others},
  booktitle={International Conference on Learning Representations},
  volume={2024},
  pages={6666--6702},
  year={2024}
}

@article{su2026generation,
  title={Generation enhances understanding in unified multimodal models via multi-representation generation},
  author={Su, Zihan and Wei, Hongyang and Cen, Kangrui and Wang, Yong and Chen, Guanhua and Yuan, Chun and Chu, Xiangxiang},
  journal={arXiv preprint arXiv:2601.21406},
  year={2026}
}

@article{yu2024representation,
  title={Representation alignment for generation: Training diffusion transformers is easier than you think},
  author={Yu, Sihyun and Kwak, Sangkyung and Jang, Huiwon and Jeong, Jongheon and Huang, Jonathan and Shin, Jinwoo and Xie, Saining},
  journal={arXiv preprint arXiv:2410.06940},
  year={2024}
}

@inproceedings{leng2025repa,
  title={Repa-e: Unlocking vae for end-to-end tuning with latent diffusion transformers},
  author={Leng, Xingjian and Singh, Jaskirat and Hou, Yunzhong and Xing, Zhenchang and Xie, Saining and Zheng, Liang},
  booktitle={2025 IEEE/CVF International Conference on Computer Vision (ICCV)},
  pages={18262--18272},
  year={2025},
  organization={IEEE}
}

@inproceedings{singh2025matters,
  title={What matters for representation alignment: Global information or spatial structure?},
  author={Singh, Jaskirat and Leng, Xingjian and Wu, Zongze and Zheng, Liang and Zhang, Richard and Shechtman, Eli and Xie, Saining},
  booktitle={The Fourteenth International Conference on Learning Representations},
  year={2025}
}

@inproceedings{xie2026unleashing,
  title={Unleashing the potential of large language models for text-to-image generation through autoregressive representation alignment},
  author={Xie, Xing and Liu, Jiawei and Lin, Ziyue and Fan, Huijie and Han, Zhi and Tang, Yandong and Qu, Liangqiong},
  booktitle={Proceedings of the AAAI Conference on Artificial Intelligence},
  volume={40},
  number={13},
  pages={11105--11113},
  year={2026}
}

@article{kim2024openvla,
  title={Openvla: An open-source vision-language-action model},
  author={Kim, Moo Jin and Pertsch, Karl and Karamcheti, Siddharth and Xiao, Ted and Balakrishna, Ashwin and Nair, Suraj and Rafailov, Rafael and Foster, Ethan and Lam, Grace and Sanketi, Pannag and others},
  journal={arXiv preprint arXiv:2406.09246},
  year={2024}
}

@article{black2024pi_0,
  title={{$\pi_0$}: A Vision-Language-Action Flow Model for General Robot Control},
  author={Black, Kevin and Brown, Noah and Driess, Danny and Esmail, Adnan and Equi, Michael and Finn, Chelsea and Fusai, Niccolo and Groom, Lachy and Hausman, Karol and Ichter, Brian and others},
  journal={arXiv preprint arXiv:2410.24164},
  year={2024}
}

@article{bjorck2025gr00t,
  title={Gr00t n1: An open foundation model for generalist humanoid robots},
  author={Bjorck, Johan and Casta{\~n}eda, Fernando and Cherniadev, Nikita and Da, Xingye and Ding, Runyu and Fan, Linxi and Fang, Yu and Fox, Dieter and Hu, Fengyuan and Huang, Spencer and others},
  journal={arXiv preprint arXiv:2503.14734},
  year={2025}
}

@article{rawal2026nord,
  title={Nord: A data-efficient vision-language-action model that drives without reasoning},
  author={Rawal, Ishaan and Gupta, Shubh and Hu, Yihan and Zhan, Wei},
  journal={arXiv preprint arXiv:2602.21172},
  year={2026}
}

@inproceedings{hu2023planning,
  title={Planning-oriented autonomous driving},
  author={Hu, Yihan and Yang, Jiazhi and Chen, Li and Li, Keyu and Sima, Chonghao and Zhu, Xizhou and Chai, Siqi and Du, Senyao and Lin, Tianwei and Wang, Wenhai and others},
  booktitle={2023 IEEE/CVF Conference on Computer Vision and Pattern Recognition (CVPR)},
  pages={17853--17862},
  year={2023},
  organization={IEEE}
}

@article{chitta2022transfuser,
  title={Transfuser: Imitation with transformer-based sensor fusion for autonomous driving},
  author={Chitta, Kashyap and Prakash, Aditya and Jaeger, Bernhard and Yu, Zehao and Renz, Katrin and Geiger, Andreas},
  journal={IEEE transactions on pattern analysis and machine intelligence},
  volume={45},
  number={11},
  pages={12878--12895},
  year={2022},
  publisher={IEEE}
}

@inproceedings{li2025enhancing,
  title={Enhancing end-to-end autonomous driving with latent world model},
  author={Li, Yingyan and Fan, Lue and He, Jiawei and Wang, Yuqi and Chen, Yuntao and Zhang, Zhaoxiang and Tan, Tieniu},
  booktitle={International Conference on Learning Representations},
  volume={2025},
  pages={42942--42959},
  year={2025}
}

@inproceedings{zhang2025epona,
  title={Epona: Autoregressive diffusion world model for autonomous driving},
  author={Zhang, Kaiwen and Tang, Zhenyu and Hu, Xiaotao and Pan, Xingang and Guo, Xiaoyang and Liu, Yuan and Huang, Jingwei and Yuan, Li and Zhang, Qian and Long, Xiao-Xiao and others},
  booktitle={2025 IEEE/CVF International Conference on Computer Vision (ICCV)},
  pages={27220--27230},
  year={2025},
  organization={IEEE}
}

@article{li2024hydra,
  title={Hydra-mdp: End-to-end multimodal planning with multi-target hydra-distillation},
  author={Li, Zhenxin and Li, Kailin and Wang, Shihao and Lan, Shiyi and Yu, Zhiding and Ji, Yishen and Li, Zhiqi and Zhu, Ziyue and Kautz, Jan and Wu, Zuxuan and others},
  journal={arXiv preprint arXiv:2406.06978},
  year={2024}
}

@inproceedings{liao2025diffusiondrive,
  title={Diffusiondrive: Truncated diffusion model for end-to-end autonomous driving},
  author={Liao, Bencheng and Chen, Shaoyu and Yin, Haoran and Jiang, Bo and Wang, Cheng and Yan, Sixu and Zhang, Xinbang and Li, Xiangyu and Zhang, Ying and Zhang, Qian and others},
  booktitle={2025 IEEE/CVF Conference on Computer Vision and Pattern Recognition (CVPR)},
  pages={12037--12047},
  year={2025},
  organization={IEEE}
}

@inproceedings{li2025end,
  title={End-to-end driving with online trajectory evaluation via bev world model},
  author={Li, Yingyan and Wang, Yuqi and Liu, Yang and He, Jiawei and Fan, Lue and Zhang, Zhaoxiang},
  booktitle={2025 IEEE/CVF International Conference on Computer Vision (ICCV)},
  pages={27137--27146},
  year={2025},
  organization={IEEE}
}

@inproceedings{xia2026drivelaw,
  title={Drivelaw: Unifying planning and video generation in a latent driving world},
  author={Xia, Tianze and Li, Yongkang and Zhou, Lijun and Yao, Jingfeng and Xiong, Kaixin and Sun, Haiyang and Wang, Bing and Ma, Kun and Chen, Guang and Ye, Hangjun and others},
  booktitle={Proceedings of the IEEE/CVF Conference on Computer Vision and Pattern Recognition},
  pages={39701--39712},
  year={2026}
}

@article{chen2024vadv2,
  title={Vadv2: End-to-end vectorized autonomous driving via probabilistic planning},
  author={Chen, Shaoyu and Jiang, Bo and Gao, Hao and Liao, Bencheng and Xu, Qing and Zhang, Qian and Huang, Chang and Liu, Wenyu and Wang, Xinggang},
  journal={arXiv preprint arXiv:2402.13243},
  year={2024}
}

@article{li2025hydra,
  title={Hydra-mdp++: Advancing end-to-end driving via expert-guided hydra-distillation},
  author={Li, Kailin and Li, Zhenxin and Lan, Shiyi and Xie, Yuan and Zhang, Zhizhong and Liu, Jiayi and Wu, Zuxuan and Yu, Zhiding and Alvarez, Jose M},
  journal={arXiv preprint arXiv:2503.12820},
  year={2025}
}

@article{shao2024deepseekmath,
  title={Deepseekmath: Pushing the limits of mathematical reasoning in open language models},
  author={Shao, Zhihong and Wang, Peiyi and Zhu, Qihao and Xu, Runxin and Song, Junxiao and Bi, Xiao and Zhang, Haowei and Zhang, Mingchuan and Li, YK and Wu, Yang and others},
  journal={arXiv preprint arXiv:2402.03300},
  year={2024}
}

@article{dauner2024navsim,
  title={Navsim: Data-driven non-reactive autonomous vehicle simulation and benchmarking},
  author={Dauner, Daniel and Hallgarten, Marcel and Li, Tianyu and Weng, Xinshuo and Huang, Zhiyu and Yang, Zetong and Li, Hongyang and Gilitschenski, Igor and Ivanovic, Boris and Pavone, Marco and others},
  journal={Advances in Neural Information Processing Systems},
  volume={37},
  pages={28706--28719},
  year={2024}
}

@article{agarwal2025cosmos,
  title={Cosmos world foundation model platform for physical ai},
  author={Agarwal, Niket and Ali, Arslan and Bala, Maciej and Balaji, Yogesh and Barker, Erik and Cai, Tiffany and Chattopadhyay, Prithvijit and Chen, Yongxin and Cui, Yin and Ding, Yifan and others},
  journal={arXiv preprint arXiv:2501.03575},
  year={2025}
}

@article{bai2025qwen25vltechnicalreport,
  title={{Qwen2.5-VL} Technical Report},
  author={Bai, Shuai and Chen, Keqin and Liu, Xuejing and Wang, Jialin
          and Ge, Wenbin and Song, Sibo and Dang, Kai and Wang, Peng
          and Wang, Shijie and Tang, Jun and others},
  journal={arXiv preprint arXiv:2502.13923},
  year={2025}
}

@misc{openscene2023,
  title={OpenScene: The Largest Up-to-Date 3D Occupancy Prediction Benchmark in Autonomous Driving},
  author={OpenScene Contributors},
  howpublished={\url{https://github.com/OpenDriveLab/OpenScene}},
  year={2023}
}

@article{cao2025pseudo,
  title={Pseudo-simulation for autonomous driving},
  author={Cao, Wei and Hallgarten, Marcel and Li, Tianyu and Dauner, Daniel and Gu, Xunjiang and Wang, Caojun and Miron, Yakov and Aiello, Marco and Li, Hongyang and Gilitschenski, Igor and others},
  journal={arXiv preprint arXiv:2506.04218},
  year={2025}
}

@article{hendrycks2016gaussian,
  title={Gaussian error linear units (gelus)},
  author={Hendrycks, Dan and Gimpel, Kevin},
  journal={arXiv preprint arXiv:1606.08415},
  year={2016}
}

\end{document}